\documentclass[letterpaper, 10 pt, conference]{ieeeconf}  

\IEEEoverridecommandlockouts                              

\usepackage{amsmath}
\usepackage{graphics}
\usepackage{graphicx}
\usepackage{amssymb}
\usepackage{color}
\usepackage{upgreek}
\usepackage{bbm, bbold}
\usepackage{algpseudocode}
\usepackage{lipsum}

\usepackage{tensor}
\usepackage[ruled,vlined]{algorithm2e}

\usepackage[normalem]{ulem}
\newcommand{\stkout}[1]{\ifmmode\text{\sout{\ensuremath{#1}}}\else\sout{#1}\fi}

\usepackage{textcomp} 

\usepackage{multicol}
\usepackage{multirow}
\usepackage{booktabs}

\usepackage[style=ieee]{biblatex}

\DeclareUnicodeCharacter{0301}{\'{e}} 

\DeclareSourcemap{
  \maps{
    \map{
      \pertype{article}
      \step[fieldset=language, null]
      \step[fieldset=url, null]
      \step[fieldset=doi, null]
      \step[fieldset=issn, null]
      \step[fieldset=isbn, null]
      \step[fieldset=note, null]
      \step[fieldset=editor, null]
      \step[fieldset=urldate, null]
      \step[fieldset=file, null]
    }
  }
}
\DeclareSourcemap{
  \maps{
    \map{
      \pertype{inproceedings}
      \step[fieldset=language, null]
      \step[fieldset=url, null]
      \step[fieldset=doi, null]
      \step[fieldset=issn, null]
      \step[fieldset=isbn, null]
      \step[fieldset=note, null]
      \step[fieldset=editor, null]
      \step[fieldset=urldate, null]
      \step[fieldset=file, null]
    }
  }
}
\DeclareSourcemap{
  \maps{
    \map{
      \pertype{incollection}
      \step[fieldset=language, null]
      \step[fieldset=url, null]
      \step[fieldset=doi, null]
      \step[fieldset=issn, null]
      \step[fieldset=isbn, null]
      \step[fieldset=note, null]
      \step[fieldset=editor, null]
      \step[fieldset=urldate, null]
      \step[fieldset=file, null]
    }
  }
}

\begin{document}

\title{\LARGE \bf 
Wake-Based Locomotion Gait Design for \textit{Aerobat}
}

\author{Eric Sihite$^{1}$ and Alireza Ramezani$^{2\,*}$%
\thanks{$^{1}$ The authors are with the Graduate Aerospace Laboratories of the California Institute of Technology (Caltech GALCIT), Pasadena, CA-91125, USA.
        {Emails: \tt\small esihite@caltech.edu}}%
\thanks{$^{2}$ The author is with the SiliconSynapse Laboratory, Department of Electrical and Computer Engineering, Northeastern University, Boston, MA-02119,USA.
    {Email: \tt\small a.ramezani@northeastern.edu}}%
\thanks{$^{*}$ This author is the corresponding author.}%
}



%

\maketitle

\begin{abstract}


Flying animals, such as bats, fly through their fluidic environment as they create air jets and form wake structures downstream of their flight path. Bats, in particular, dynamically morph their highly flexible and dexterous armwing to manipulate their fluidic environment which is key to their agility and flight efficiency. 
This paper presents the theoretical and numerical analysis of the wake-structure-based gait design inspired by bat flight for flapping robots using the notion of reduced-order models and unsteady aerodynamic model incorporating Wagner function. The objective of this paper is to introduce the notion of gait design for flapping robots by systematically searching the design space in the context of optimization. The solution found using our gait design framework was used to design and test a flapping robot.  
\end{abstract}

\IEEEpeerreviewmaketitle

\definecolor{green}{rgb}{0.96, 0.29, 0.54}

\section{Introduction}
\label{sec:intro}

When a flapping bat propels through its fluidic environment, it creates periodic air jets in the form of wake structures in the downstream of its flight path. The animal's notable dexterity to quickly manipulate these wakes with fine-grained, fast body adjustments is key to retaining the force-moment needed for an all-time controllable flight, even near stall conditions, sharp turns, and heel-above-head maneuvers. We refer to bats' locomotion based on dexterously manipulating the fluidic environment through dynamically versatile wing conformations as \textit{dynamic morphing wing flight}. 

Despite the known ability of flying vertebrates such as bat in dynamically adjusting their wing planform configuration, existing bioinspired micro aerial vehicles (MAVs) cannot match these biological systems. Dynamic morphing is key to bats efficiency and agility.

The research gap in copying vertebrates flight is evident in two domains of robot design and locomotion control. As it can be found out by inspecting current literature, a large number of flapping robot designs have been introduced, including small \cite{farrell_helbling_review_2018, phan_insect-inspired_2019, madangopal_biologically_2005, dhingra_device_2020, rosen_development_2016, richter_untethered_2011, phan_towards_2020, zhang_design_2017, ma_controlled_2013, ristroph_stable_2014, festo_team_bionicopter_nodate, noauthor_artifacts_nodate} and medium to larger robots \cite{gerdes_robo_2014, tedrake_learning_2009, de_croon_design_2009, tu_untethered_2020, holness_design_2019, grauer_testing_2011, lee_experimental_2012, rose_comparison_2014, yang_dove_2018, peterson_wing-assisted_2011, peterson_experimental_2011}, which all share a common design theme of insect- or hummingbird-style designs. In all of these systems, the wings are widely made of a single segment, without wing planform articulations. The wings can translate rapidly relative to the body \cite{madangopal_biologically_2005, yang_dove_2018, sane_lift_2001}. Hence, these systems are technically scaled up insect robots. 

Conversely, the MAVs with adaptive structures \cite{chang_soft_2020, di_luca_bioinspired_2017, abdulrahim_flight_2004, basaeri_experimental_2014, noauthor_roboswift_nodate, daler_bioinspired_2015} possess articulated wings. Some of them have combined standard fixed- or rotary-wing designs with bioinspiration from birds \cite{chang_soft_2020, di_luca_bioinspired_2017, abdulrahim_flight_2004, basaeri_experimental_2014, noauthor_roboswift_nodate}. These systems are not flapping machines, possess morphing wings, and can move the wings in multiple directions including sweeping, mediolateral motions, and flexion-extensions. However, despite their adaptive wing structures, we do not refer to these examples as dynamic morphing wings because they cannot achieve dynamic wing planform conformations. Instead, they adjust wing structures in a quasi-static fashion, i.e., slow adjustments whose time duration do not fit to a gaitcycle (roughly one-tenth of a second).

\begin{figure}[t!]
    \vspace{0.1in}
    \centering
    \includegraphics[width=1.0\linewidth]{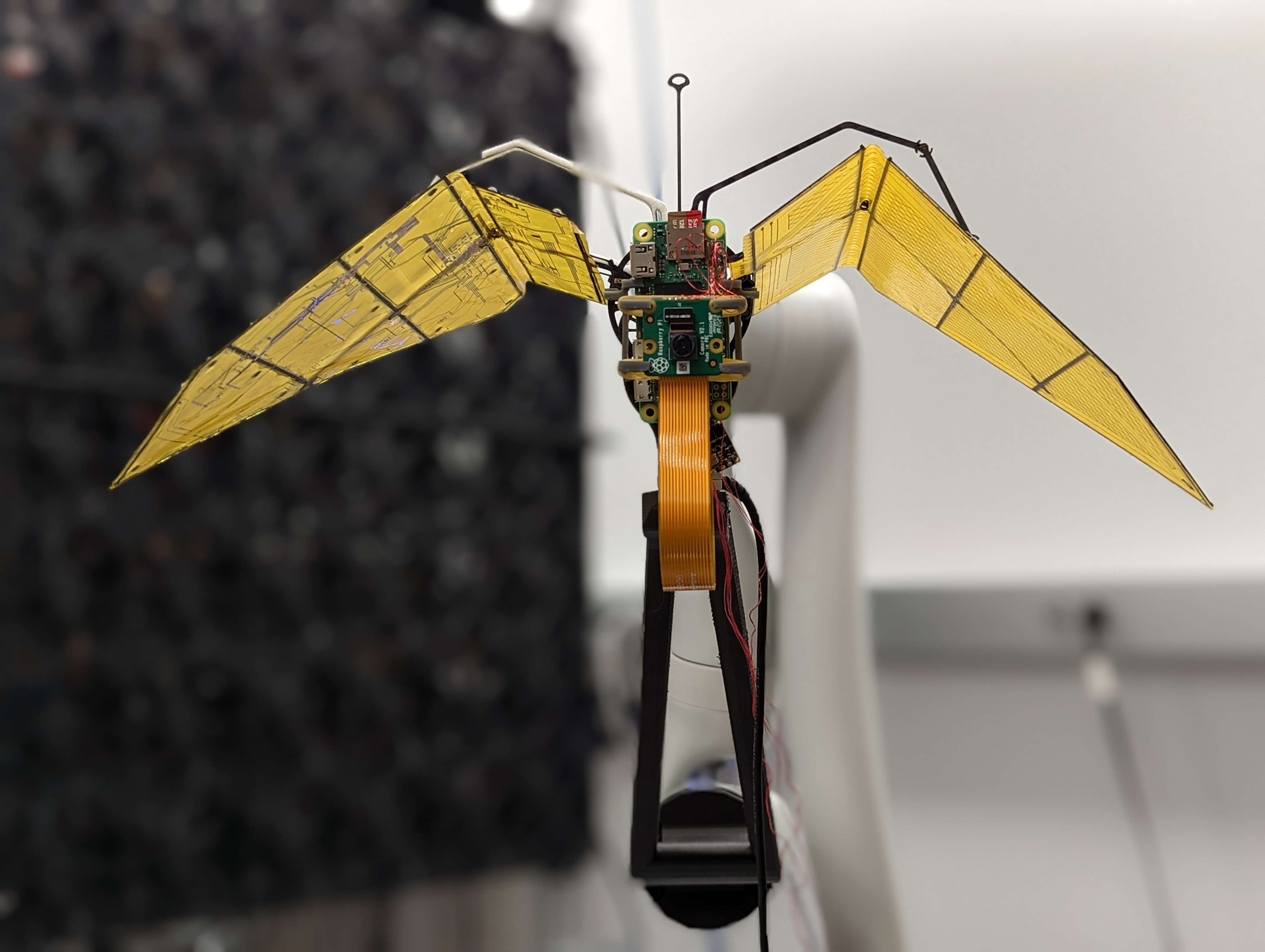}
    \caption{Shows Northeastern University's morphing wing robot, \textit{Aerobat}.}
    \label{fig:cover}
    \vspace{-0.1in}
\end{figure}

The other research gap is that notions such as locomotion gait (i.e., time-evolution of joint trajectories) and posture control (i.e., regulation of locomotion gait using closed-loop feedback) are completely overlooked in flapping wing flight. This paper aims at making contributions in this area. Specifically speaking, we introduce aerial locomotion gait for bioinspired flapping MAV. Locomotion gait has no clear definition for aerial robots. Currently, insect- or hummingbird-style models dominating the body of bioinspired aerial robot works assume no locomotion gait design or posture control while in biology flying animals have been studied based on their aerial gaits for long times. 

Bat aerial gait is often characterized based on topological wake structures that bats generate. Bats are known for the complex wake structures that they can generate. This wake-structure-based gait definition differs from that in terrestrial systems which is based on joint trajectories. 

From a gait design standpoint, a wake-structure-based view towards gait design possesses a few merits. First, it is biologically meaningful. Second, it allows to shift gait parameterization from joint space, which can be high dimensional, to 3D Euclidean space yielding less costly designs. 

The objective of this paper is to introduce the notion of gait design for flapping robots. Of course, in fluidic-based locomotion, the wake structure not only depends on joint motions but also the morphology, shape and geometry of body. As a result, our gait design approach yields a systematic search method in the robot morphology space.

Briefly speaking, a desired gait taken from bat study literature is considered \cite{hubel_wake_2010}. Then, based on the notion of reduced-order models (zero-dynamics) borrowed from legged locomotion and inexpensive geometric fluid dynamics robot designs are found to match the desired gait. While our proposed technique to match a gait is fully numerical and misses experimental validation approaches such as particle image velocimetry, the overall gait design approach remains valid as the numerical wake structure predictions can always be improved with more accurate indicial models.

We designed a flapping MAV, shown in Fig.~\ref{fig:cover}, with dynamic wing planform articulations based on the proposed design method and tested the feasibility of flight in this system \cite{sihite_computational_2020, sihite_enforcing_2020, sihite_integrated_2021, sihite2022unsteady, sihite2021orientation, lessieur2021mechanical, ramezani_biomimetic_2017, ramezani_bat_2016, ramezani_modeling_2016, ramezani_describing_2017}. In our real-world tests, the wing planform articulations required to generate the desired wake structure was achieved using embodied morphing, a concept we introduced in our previous works. In embodied morphing mechanical intelligence is employed.
   
\section{Dynamic Morphing Wing Flight Model}
\label{sec:model}

Modeling dynamic morphing wing flight involves a high dimensional shape space that make parametric gait design hard, if not impossible. To combat this challenge we use the notion of zero-dynamics and 3D wake-structures. 
For this purpose we consider a cascade dynamical model.

In the cascade model proposed here, the inertial and aerodynamics contributions are interconnected through the computational roles prescribed by closed-loop feedback or morphology. We denote these computational roles with the holonomic constraint $y_1=h_1(q_a)$, where $q_a$ is the vector of shape variables. 

The cascade model is given by
\begin{equation}
\begin{aligned}
\Sigma_{Full}&:\left\{
\begin{aligned}
    \dot x &= f(x) + g_1(x) u + g_2(x) y_2 \\
    y_1 &= h_1(x)\\
\end{aligned}
\right.\\
\Sigma_{Aero}&:\left\{
\begin{aligned}
    \dot \xi &= A_\xi (t) \xi + B_\xi (t) y_1 \\
    y_2 &= C_\xi (t)\xi + D_\xi (t) y_1
\end{aligned}
\right.
\end{aligned}
\label{eq:ss-rep-fulldyn}
\end{equation}
\noindent where $t$, $x$, and $\xi$ denote time, the state vector and aerodynamic hidden variables, respectively. In \eqref{eq:ss-rep-fulldyn}, $f(x)$ and $g_1(x)$ embody the inertial terms from Euler-Lagrange formalism, the aerodynamic force output $y_2$ gives the instantaneous external forces whose dynamics are captured in the state-space form by $A_\xi$, $B_\xi$, $C_\xi$, and $D_\xi$ matrices which will be explained in section~\ref{sec:wake}. 


Briefly speaking the state vector $x$ is defined such that it yields a straightforward form for the internal dynamics. However, more complex internal models that capture aeroelasticity could be included. Let $q_u = [q_{E},p_{b}]^\top$ be the underactuated (passive) coordinates where the orientations and body positions are parameterized with the Euler angles $q_{E}$ and 3D position vector $p_{b}$. And, let the shape variable vector $q_a$ denote the actuated (active) coordinate vectors, i.e., dynamic morphing wing joint angles. These underactuated and actuated coordinates result in a system that can be fully determined by the configuration variable vector $q = [q_u^\top, q_a^\top]^\top$ and the state vector $x = [q^\top, \dot q^\top]^\top$. 

As it was mentioned before, in dynamic morphing wing flight, the dimensionality of $q_a$ is very high and $\dot q_a$ takes large values yielding infeasible gait and robot design problems. So, next, we will use the notion of restriction dynamics of \eqref{eq:ss-rep-fulldyn} and 3D geometric representation of wake structures to design gait for our system.

\section{Zero Dynamics}

In \eqref{eq:ss-rep-fulldyn}, we can employ the primer parameter $\omega(t)$ to regulate $y_2$ in a quasi-static fashion around pre-defined equilibrium. This concept is inspired by reference governor method from control literature. Consider the pre-stabilized ($u$ used to enforce $y_1(x,\omega)$) version of \eqref{eq:ss-rep-fulldyn}
\begin{equation}
    \begin{aligned}
        \dot{x}_z &= f_z(x_z)+g_z(x_z)\omega(t)\\
    \end{aligned}
    \label{eq:RG-model-simplified}
\end{equation}
\noindent where $x_z=[q^\top_u,\dot q^\top_u]$. It is assumed the dynamic state feedback is successfully designed and we skip further details since it is not core to our discussions. As a result, $g_1(x)u$ in \eqref{eq:RG-model-simplified} is re-written in a way that allows an affine-in-primer form given by \eqref{eq:RG-model-simplified}. Other nonlinear terms that are only state-dependent are summarized under the nonlinear offset term $f_z(x_z)$, which denotes the restriction dynamics over the zero-dynamics manifold. Note that $\mathrm{dim}\{x_z\}<<\mathrm{dim}\{x\}$. This property of \eqref{eq:RG-model-simplified} is one of the properties that we exploit for gait design with an optimizer. The next property is the low-dimensional representation of gait parameterization in the 3D space which follows.

\section{Wake Structure Prediction Based on Horseshoe Vortex Method}
\label{sec:wake}

Gait parameterization in its basic form assumes polynomials for each shape variables in $q_a$. These parameters are tweaked by an optimizer to minimize a performance index. This view is broadly used in terrestrial systems and its application in dynamic morphing wing flight can yield costly optimization. Therefore, we are interested to inspect a new form of gait design that not only is biologically meaningful but also reduces cost of our computations. 3D wake structures behind the flapping wings can serve our objectives nicely.

We briefly explain how wake-structure-based gaits are efficiently computed. Our method is based on closed-form solutions of the unsteady lifting line theory incorporating Wagner's function to capture transient 3D flow behavior, as derived in fluid dynamics textbooks and recent publications \cite{izraelevitz_state-space_2017,boutet_unsteady_2018}. We superimpose horseshoe vortices on and behind the wing blade elements to visualize the wake structures and use them as 3D geometries to design gaits.

The time-varying circulation value at $s_i$, the location of the i-th element on the wing, is denoted by $\Gamma_i(t)$ and is parameterized by truncated Fourier series of $n$ coefficients. The number of Fourier coefficients are equal to the number of wing segments to ensure the existence of unique solutions. $\Gamma_i$ is given by

\begin{equation}
\begin{aligned}
    \Gamma_i(t) = a^\top(t)
    \begin{bmatrix}
    \sin(\theta_i)\\
    \vdots\\
    \sin(n\theta_i)\\
    \end{bmatrix},
\end{aligned}
    \label{eq:circulation}
\end{equation}
\noindent where $a=[a_1,\dots,a_n]^\top$ are the Fourier coefficients and $\theta_i=\arccos(\frac{s_i}{l})$ ($l$ is the wingspan size). From Prandtl's lifting line theory, an additional circulation-induced kinematics denoted by $y_{\Gamma}$ are considered on all of the quarter-chord points $p_i$. These circulation-induced kinematics are given by

\begin{equation}
\begin{aligned}
    y_{\Gamma} = 
    \begin{bmatrix}
    1&\frac{\sin{2\theta_1}}{\sin{\theta_1}}&\dots&\frac{\sin{n\theta_1}}{\sin{\theta_1}}\\
    1&\frac{\sin{2\theta_2}}{\sin{\theta_2}}&\dots&\frac{\sin{n\theta_2}}{\sin{\theta_2}}\\
    \vdots\\
    1&\frac{\sin{2\theta_n}}{\sin{\theta_n}}&\dots&\frac{\sin{n\theta_n}}{\sin{\theta_n}}\\
    \end{bmatrix}a(t)
    \label{eq:induced-kin}
\end{aligned}
\end{equation}

We use a Wagner function $\Phi(\tau) = \Sigma_{k=1}^2\psi_k \exp{(-\frac{\epsilon_k}{c_i} \tau)}$, where $\psi_k$, $\epsilon_k$ are some scalar coefficients and $\tau$ is a scaled time to compute the aerodynamic force coefficient response $\beta_{i}$ associated with the i-th blade element. The kinematics of i-th element using Eqs.~\ref{eq:ss-rep-fulldyn} and \ref{eq:induced-kin} are given by 
\begin{equation}
y'_{1,i} = y_{1,i} + y_{\Gamma,i}.    
\end{equation}

\noindent Following Duhamel's integral rule, the response is obtained by the convolution integral given by

\begin{equation}
    \beta_i = y'_{1,i}\Phi_0 + \int_0^t\frac{\partial \Phi(t-\tau)}{\partial \tau}y'_{1,i}d\tau \\
\label{eq:lift_coeff_wagner}
\end{equation}

\noindent where $\Phi_0=\Phi(0)$. We perform integration by part to eliminate $\frac{\partial \Phi}{\partial \tau}$, substitute the Wagner function given above in \eqref{eq:lift_coeff_wagner}, and employ the change of variable given by $z_{k,i} (t) = \int_{0}^{t} \exp{(-\frac{\epsilon_k}{c_i}(t-\tau))} y'_{1,i} d\tau$, where $k\in\{1,2\}$, to obtain a new expression for $\beta_i$ based on $z_{k,i}$  

\begin{equation}
\beta_i =
y'_{1,i}\Phi_0+
\begin{bmatrix}
\psi_1\frac{\epsilon_1}{c_i}&\psi_2\frac{\epsilon_2}{c_i}
\end{bmatrix}
\begin{bmatrix}
z_{1,i}\\
z_{2,i}
\end{bmatrix}
\label{eq:?}
\end{equation}

The variables $a$, $z_{1,i}$, and $z_{2,i}$ are used towards obtaining a state-space realization that can be marched forward in time. Using the Leibniz integral rule for differentiation under the integral sign, unsteady Kutta–Joukowski results $\beta_i=\frac{\Gamma_i}{c_i}+\frac{d\Gamma_i}{dt}$, and \eqref{eq:circulation}, the model that describes the time evolution of the aerodynamic states is obtained

\begin{equation}
\Sigma_{Aero,i}:\left\{
\begin{aligned}
    A_i\dot{a} &= -B_ia + C_iZ_i  + \Phi_0y'_{1,i}\\
    \dot{Z}_i &= D_iZ_i + E_iy'_{1,i}\\
\end{aligned}
\right.
    \label{eq:aerodynamic-ss-model}
\end{equation}


\noindent where $Z_i$, $A_i$, $B_i$, $C_i$, $D_i$, and $E_i$ are the state variable and matrices corresponding to the i-th blade element. They are given by

\begin{equation}
\begin{aligned}
    Z_i &= 
    \begin{bmatrix}
    z_{1,i} & z_{2,i}
    \end{bmatrix}^\top,\\
    A_i &= 
    \begin{bmatrix}
    \sin{\theta_i} & \sin{2\theta_i} & \dots & \sin{n\theta_i} 
    \end{bmatrix},\\
    B_i &= A_i/c_i,\\
    C_i &= 
    \begin{bmatrix}
    \frac{\psi_1\epsilon_1}{c_i} & \frac{\psi_2\epsilon_2}{c_i}
    \end{bmatrix},\\
    D_i &= 
    \begin{bmatrix}
    \frac{-2\epsilon_1}{c_i} & 0\\
    0 & \frac{-2\epsilon_2}{c_i}
    \end{bmatrix},\\
    E_i &=
    \begin{bmatrix}
    2-\exp{\frac{\epsilon_1 t}{c_i}} & 2-\exp{\frac{\epsilon_2 t}{c_i}}
    \end{bmatrix}^\top.
\end{aligned}
    \label{eq:aerodynamic-ss-model-elements}
\end{equation}

\begin{figure}[t]
\vspace{0.1in}
    \centering
    \includegraphics[width=0.7\linewidth]{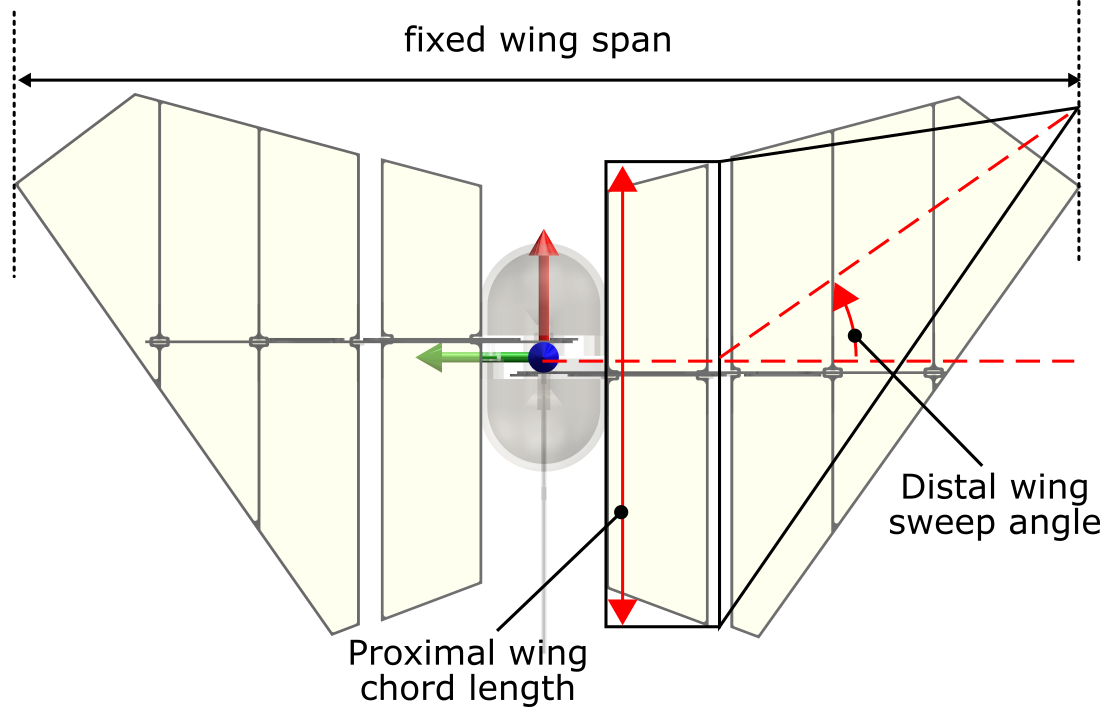}
    \caption{Wing parameters}
    \label{fig:morphology-params}
\vspace{-0.1in}
\end{figure}

\noindent We define the unified aerodynamic state vector, used to describe the state space of \eqref{eq:ss-rep-fulldyn}, as following $\xi = [a^\top, Z^\top]^\top$, where $Z=[Z^\top_1,\dots,Z^\top_n]^\top$. The state, control, and output matrices used to describe the unified aerodynamic system $\Sigma_{Aero}$ are obtained by stacking the matrices from \eqref{eq:aerodynamic-ss-model-elements} as following 
\begin{equation}
\begin{aligned}
 A &= 
\begin{bmatrix}
A^\top_1 & \dots & A^\top_n\\
\end{bmatrix}^\top\\
C &=
\begin{bmatrix}
C_1 & 0 & \dots & 0\\
0 & C_2 & \dots & 0\\
\vdots & & \ddots & \vdots\\
0 & \dots & 0 & C_n\\
\end{bmatrix}
\end{aligned}
\end{equation}
\noindent where $B$ and $E$ are obtained similar to $A$. The matrix $D$ is found similar to $C$ from the equation above. Note that since the number of blade elements and truncated Fourier coefficients are equal, the matrix $A^{-1}$ is well-defined. Consequently, the terms in \eqref{eq:ss-rep-fulldyn} are given by 
\begin{equation}
\begin{aligned}
 A_\xi &= \begin{bmatrix}
A^{-1}&0\\0&I
\end{bmatrix}
\begin{bmatrix}
-B&C\\0&D
\end{bmatrix},\\
B_\xi &= \begin{bmatrix}
A^{-1}&0\\0&I
\end{bmatrix}   
\begin{bmatrix}
\Phi_0I\\E
\end{bmatrix},
\end{aligned}
\end{equation}
\noindent $C_\xi=C$, and $D_\xi=\Phi_0I$ where $I$ is the identity matrix. 

According to Kutta-Joukowski (\eqref{eq:aerodynamic-ss-model}) the unobservable state vector $\xi$ can be parameterized or regulated by the output dynamics $y_1$. In other words, based on \eqref{eq:circulation}, $y_1$ can be used to servo the circulation distribution through the state vector $\xi$. Hence, the wake structure defined by the circulation distribution using the horseshoe method can be assumed as the gait on the zero-dynamics manifold.

One last step remains to be taken before morphology response ($y_1=h(x)$) and gait can be linked. A set of $m$ wake points $w_i$ are initially placed at equal distances with respect to each other in the wake region behind the dynamic morphing wings. The time-varying circulation distribution $\Gamma(t)$ and its induced kinematics at the wake points are obtained using Biot-Savart theorem. 

We use the wake points' kinematics to obtain the vertices, edges, and faces (mesh faces are initially quadrilateral) of the polygon meshes that define the shape of a 3D geometry. We refer to this 3D geometry as the wake structure denoted by $\mathcal{W}$. For the reason that Biot-Savart theorem is core to our derivations, we introduce a map called Biot-Savart map, denoted by 
\begin{equation}
    \mathcal{W}=\mathcal{B}(x_z,\xi,\dot \xi).
    \label{eq:biot-savart}
\end{equation}
\noindent This map succinctly captures all steps taken above to generate the topological wake structures based on feedback or morphology roles, $y_1$. 

The existence of $\mathcal{B}(x_z,\xi,\dot \xi)$ shown above is important because it demonstrates that any conceivable wake structures (gaits) are possible as long as any $y_1$ is enforceable. Therefore, it is important to demonstrate that both closed-loop feedback or embodiment can be used for this purpose. In our past works, we have shown that embodiment, the use of mechanical intelligence, can be a reliable option to enforce $y_1$ when many active DOFs are involved. The experimental results presented in Section~\ref{sec:results} uses mechanical intelligence to realize the desired gaits.

\section{Wake-Structure-Based Gait Design}
\label{sec:gait-design}

The gait design optimization problem aims at identifying a robot morphology that yields a desired gait, that is, a desired wake structure. This optimization problem is given by

\begin{equation}
\mathfrak{M}:~\left\{
\begin{aligned}
\underset{\mathcal{Z}}{\textbf{\textrm{min}}} \quad & (\mathcal{W}-\mathcal{W}_{d})^\top(\mathcal{W}-\mathcal{W}_{d})\\
\textbf{\textrm{s.t.}} \quad & \dot{x}_z - f_z(x_z) = 0\\
    & \mathcal{W}-\mathcal{B}(x_z,\dot \xi) = 0\\
\end{aligned}\right.
\label{eq:comp_struct_opt}
\end{equation}

\noindent where $\mathcal{Z}=\{x|y_1(x)=0,~L_{f}y_1(x)=0\}$ denotes the zero-dynamics manifold. In \eqref{eq:comp_struct_opt}, the cost function, the euclidean distance between the actual $\mathcal{W}$ and desired $\mathcal{W}_d$ wake points, is minimized across the zero-dynamics manifold. The indicial model derived previously and the Biot-Savart map are considered as part of the constraints. They are used to determine the wakes during one gaitcycle in terms of the coordinates from $\mathcal{Z}$. 

We note that based on how the zero-dynamics manifold is defined above, the wake-structure-based gait design heavily depends on the form of the output function $y_1=h(x)$. For example, see the result section were the output functions corresponding to the two- and three-axes morphing systems yield different gaits.

In addition, part of the gait design occur through physical property tweaking. Meaning, besides adjusting $y_1=h(x)$, this step involves adjusting the size, material properties, and wing surface geometry. We briefly describe how physical properties are parameterized. 

We parameterize the space of wing shapes across two dimensions, chord length and sweeping angle according to Fig.~\ref{fig:morphology-params}. We constrained our design space to keep proximal and distal wing spans constant so that the wing is compatible with $y_1=h(x)$. As a result, the design space of this wing can be parameterized with two components: the proximal wing chord length and the distal wing sweeping angle. Next, we briefly discuss our designed gaits and report the prototype we built to realize these gaits.

\section{Results}
\label{sec:results}

\begin{figure}[t]
\vspace{0.1in}
    \centering
    \includegraphics[width=1.0\linewidth]{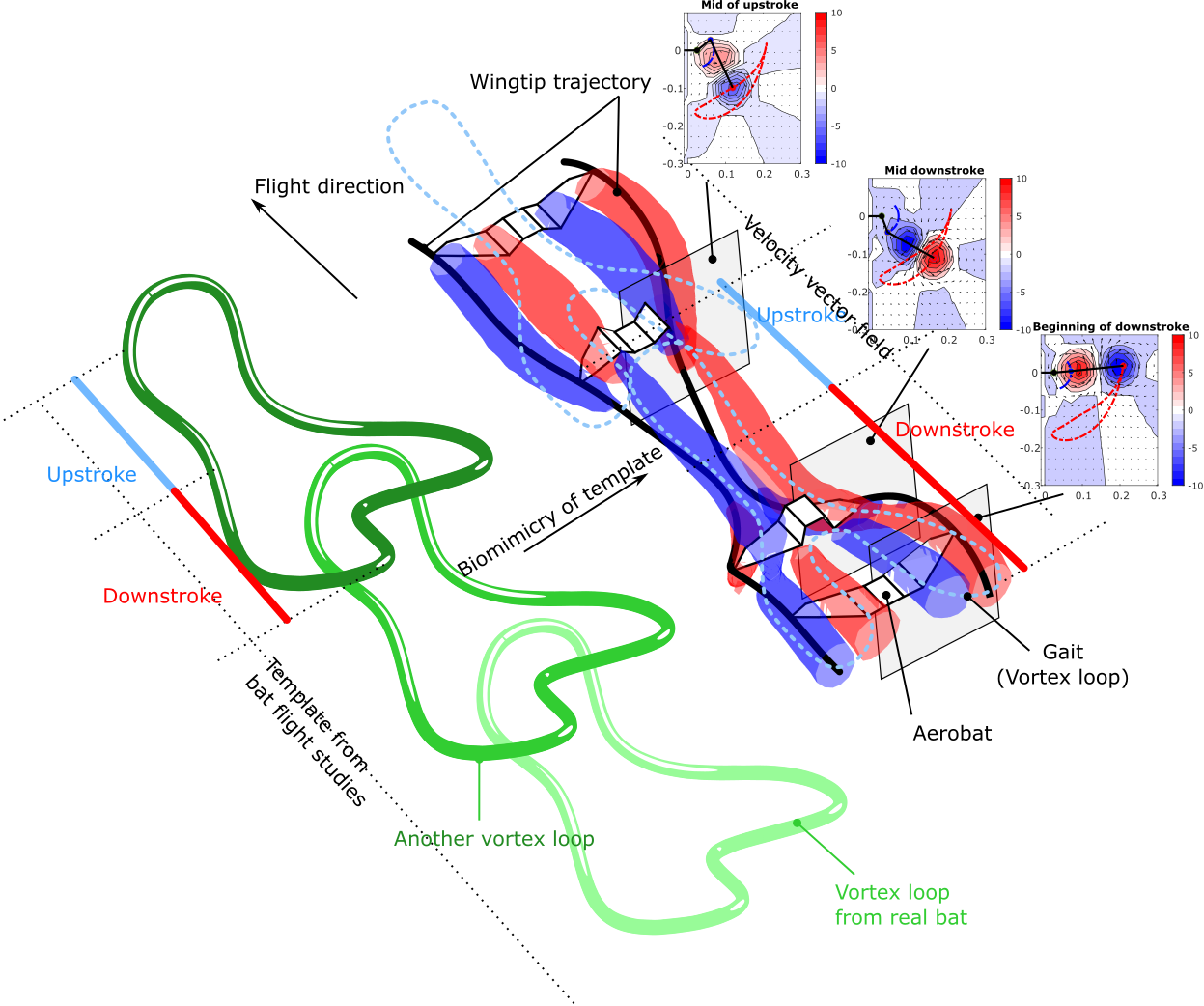}
    \caption{A comparison between gait generated by our morphing wing robot, Aerobat, and real bat \cite{hubel_wake_2010}.}
    \label{fig:snapshot}
\vspace{-0.1in}
\end{figure}

\begin{figure}[t]
\vspace{0.1in}
    \centering
    \includegraphics[width=\linewidth]{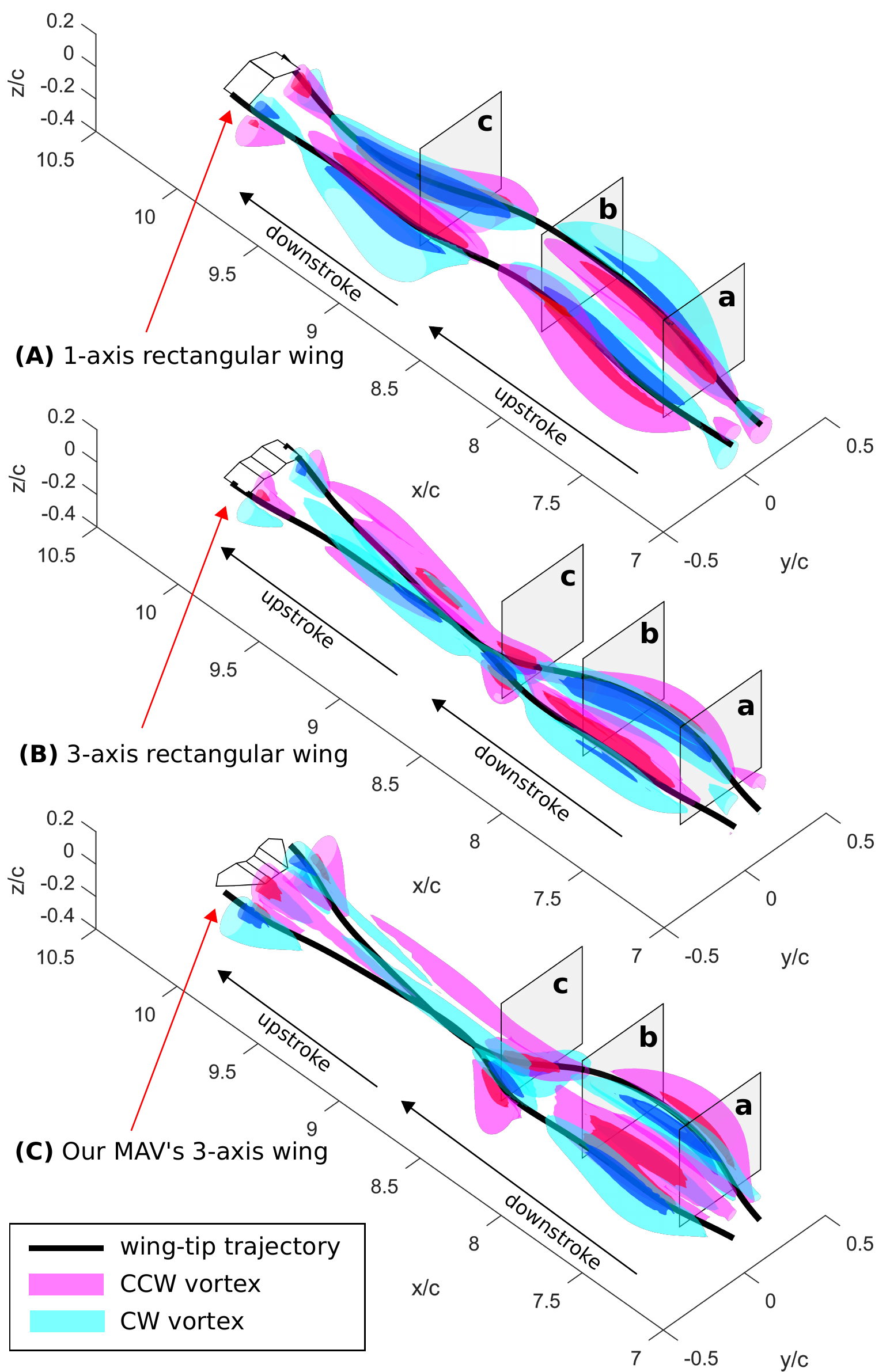}
    \caption{Iso-surface plots showing the vortices generated by the flapping wing across different output functions $y_1=h(x)$, i.e., two- and three-axes dynamic morphing wing styles, under the same conditions. The vortex is defined about the $x$-axis (forward direction) and the iso-surfaces are plotted with the following thresholds for red, magenta, blue, and cyan, respectively: 300, 100, -300, -100.}
    \label{fig:vortex_plots}
\vspace{-0.1in}
\end{figure}

\begin{figure}[t]
\vspace{0.1in}
    \centering
    \includegraphics[width=\linewidth]{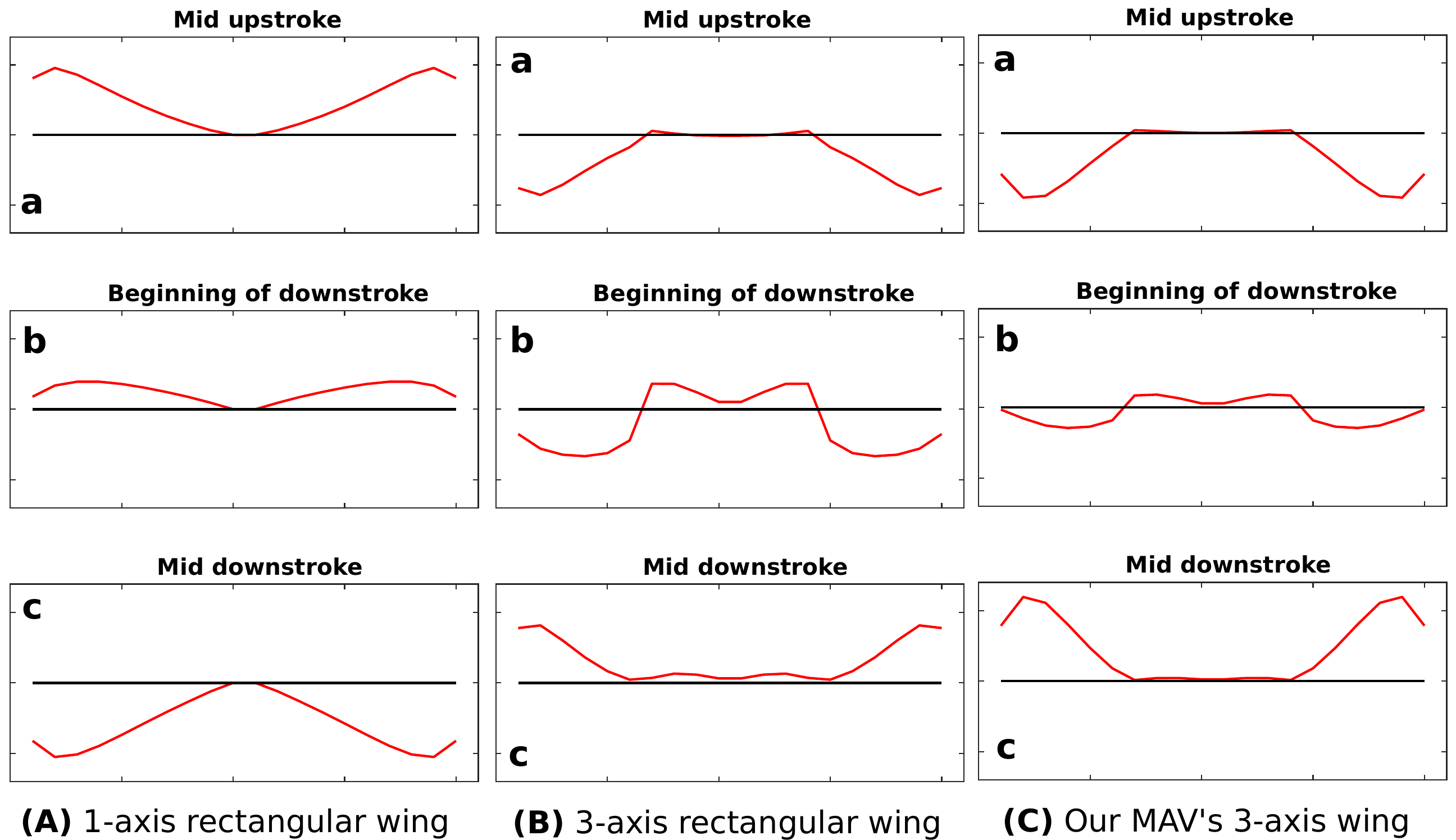}
    \caption{Circulation distribution function across the entire wing surface at the specified time as shown in Fig. \ref{fig:vortex_plots}.}
    \label{fig:gamma_plots}
\vspace{-0.1in}
\end{figure}

\begin{figure}[t]
\vspace{0.1in}
    \centering
    \includegraphics[width=\linewidth]{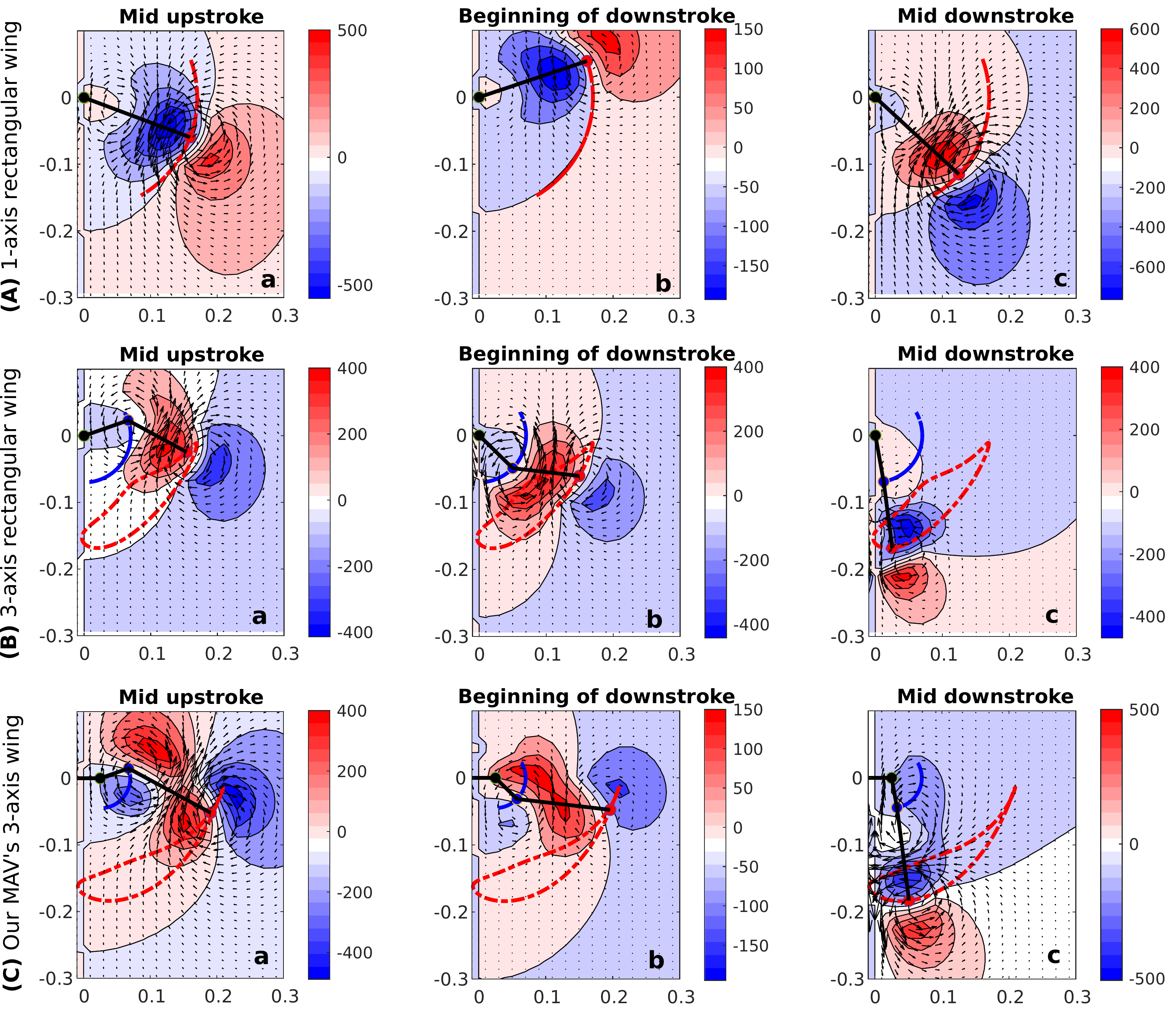}
    \caption{Strength and velocity vector field of the induced vortices at the proximity of the left wing, defined about the $x$-axis. The plots are evaluated at the specified time shown in Fig. \ref{fig:vortex_plots}.}
    \label{fig:sectional_vortex_plots}
\vspace{-0.1in}
\end{figure}

We numerically searched for output functions, $y_1=h(x)$, and wing shapes that satisfy our desired gait design. The output function defines the nonlinear relationship between wing shape variables and determines morphing behavior. We considered two forms where two (one-axis) or four (three-axes) shape variables are involved. 

We developed our simulator in Matlab using the zero dynamics and indicial models described above in Sections~\ref{sec:model} and \ref{sec:wake} to generate and study gait design for flapping MAV. The simulation produces the magnitude and direction of the horseshoe vortex on each blade elements on the wing, which can be used to calculate the vortex-induced velocity flow field behind the trailing edge.

The gait design was performed subjected to the same upstream air flow conditions, flapping speed, and total wing span size (i.e., tip-to-tip length). It is assumed that the robot is constrained to moving at a constant forward speed of 1 m/s with a flapping frequency of 2 Hz. All wings have a mean chord length of 15 cm and a wingspan of 34 cm. The three-axes rectangle wing and our robot's design share the same output function. However, the wing shape in our design is tweaked according to the parameterization convention explained before to achieve the desired gait. 

The simulation results can be seen in Fig. \ref{fig:vortex_plots} to \ref{fig:sectional_vortex_plots}. Figure \ref{fig:vortex_plots} shows the iso-surface plots of the induced vortex as the robot moves forward along the $x$-axis. We then evaluate the generated planar vortices at three time sections shown in Fig. \ref{fig:vortex_plots} and labeled using lowercase letters to further study velocity vector fields. These vector fields can reveal information regarding jet formation which determine lift and drag force generation.   

Figure \ref{fig:gamma_plots} shows the magnitude and direction of the horseshoe vortices on the wings' blade elements, while Fig. \ref{fig:sectional_vortex_plots} shows the induced vortex strength and the velocity vector field at the cross-sections shown in Fig. \ref{fig:vortex_plots}.

As shown in Fig. \ref{fig:vortex_plots}, the one-axis wing design has a significantly larger upstroke vortices compared to the other designs, where our robot's three-axes design has the smallest upstroke vortices. The three-axes design captures the wing folding during the upstroke motion which reduces the induced vortex and the negative lift, resulting in a more efficient flapping gait. 

Additionally, the vortices generated by our MAV's wing design are smaller during upstroke and larger during the downstroke compared to the rectangular wing design. We prototyped the design and the snapshots of the flight tests using an external power supply is shown in Fig.~\ref{fig:snapshot}. 

While we lack the experimental results to show the wake structure generated by the prototype match or do not match the numerical results reported here, the framework proposed here views bioinspired flapping robot design from a fundamentally different perspective. Our intention is to verify the accuracy of the robot design method by the time-resolved capture of wake structures in a wind tunnel.

\section{Conclusion} 
\label{sec:conclusion}

In this paper, we presented a systematic framework to design gait for flapping MAVs. The wake structures generated by the system form 3D geometries. We refer to these 3D geometries as aerial locomotion gait. This definition of gait is inspired by bat studies. We outlined our theoretical gait design framework based on the notion of zero dynamics of flight and unsteady aerodynamic horseshoe vortices which yield the wake structures. We designed a robot based on the desired gait. In our future work, we look to perform experimental validation of our wake structure prediction and gait design framework using methods such as particle image velocimetry.

\printbibliography

\end{document}